\documentclass[letterpaper, 10 pt, conference]{ieeeconf}  % Comment this line out if you need a4paper

\IEEEoverridecommandlockouts                              % This command is only needed if 
\usepackage[normalem]{ulem}
\usepackage{subcaption}
\usepackage{algorithm}
\usepackage{algpseudocode}
\usepackage{amssymb}  
\usepackage{amsfonts}       % blackboard math symbols
\usepackage{amsmath}
\usepackage{mathtools}
\usepackage{bm}
\usepackage{bbm}
\usepackage{changepage}
\usepackage[makeroom]{cancel}
\usepackage{sidecap}
\usepackage{longtable}
\usepackage{pifont}
\usepackage[dvipsnames]{xcolor}
\usepackage{textcomp,    % for \textlangle and \textrangle macros
	xspace}
\usepackage{pgfplots}
\usepackage{tikz} 
\usepackage{bigints}
\usepackage{xcolor} 
\usepackage{colortbl}  
\usepackage{multirow}
\usepackage{adjustbox}
\usepackage{url}
\usepackage{cite}

\usetikzlibrary{pgfplots.groupplots}
\usepackage{hyperref}
\newcommand*\de{\mathop{}\!\mathrm{d}}

\DeclareMathOperator*{\argmax}{arg\max}

\newcommand{\muvec}{\bm{\mu}}

\newcommand{\epsvec}[0]{\bm{\epsilon}}

\newcommand{\zvec}{\mathbf{z}}

\newcommand{\Bvec}{\mathbf{B}}
\newcommand{\Cvec}{\mathbf{C}}

\newcommand{\Sigmavec}{\bm{\Sigma}}

\newcommand{\cvec}{\mathbf{c}}

\newcommand{\omegavec}{\bm{\omega}}
\newcommand{\pivec}{\bm{\pi}}
\newcommand{\Omegavec}{\bm{\Omega}}

\usepackage{todonotes}
\newcounter{todocounter}

\title{\LARGE \bf
Contextual Latent-Movements Off-Policy Optimization\\ for Robotic Manipulation Skills}

\author{Samuele Tosatto$^{1}$,
        Georgia Chalvatzaki$^{1}$ and Jan Peters$^{1}$
\thanks{$^1$Intelligent Autonomous Systems, Technische Universit\"{a}t Darmstadt, Hochschulstr. 10, 64289 Darmstadt, Germany \tt \small  \{samuele, georgia\}@robot-learning.de}
\thanks{This research is financially supported by the Bosch-Forschungsstiftung program and the EU Horizon 2020 research and innovation program under grant agreement \#640554 (SKILLS4ROBOTS).}}
\begin{document}

\maketitle
\thispagestyle{empty}
\pagestyle{empty}

%%%%%%%%%%%%%%%%%%%%%%%%%%%%%%%%%%%%%%%%%%%%%%%%%%%%%%%%%%%%%%%%%%%%%%%%%%%%%%%%
\begin{abstract}
% What is the problem
Parameterized movement primitives have been extensively used for imitation learning of robotic tasks. However, the high-dimensionality of the parameter space hinders the improvement of such primitives in the reinforcement learning (RL) setting, especially for learning with physical robots. In this paper we propose a novel view on handling the demonstrated trajectories for acquiring low-dimensional, non-linear latent dynamics, using mixtures of probabilistic principal component analyzers (MPPCA) on the movements' parameter space. Moreover, we introduce a new contextual off-policy RL algorithm, named LAtent-Movements Policy Optimization (LAMPO). LAMPO can provide gradient estimates from previous experience using self-normalized importance sampling, hence, making full use of samples collected in previous learning iterations. These advantages combined provide a complete framework for sample-efficient off-policy optimization of movement primitives for robot learning of high-dimensional manipulation skills. Our experimental results conducted both in simulation and on a real robot show that LAMPO provides sample-efficient policies against common approaches in literature. Code available at \url{https://github.com/SamuelePolimi/lampo}.

\end{abstract}

%%%%%%%%%%%%%%%%%%%%%%%%%%%%%%%%%%%%%%%%%%%%%%%%%%%%%%%%%%%%%%%%%%%%%%%%%%%%%%%%
\section{INTRODUCTION}

Learning manipulation skills is essential for enabling robots to execute many tasks, both in-home and industrial environments. An essential aspect of future robots is their ability to acquire, adapt or improve their skills, using demonstrations from non-expert users \cite{nonexperts}. The provision of demonstrations is essential for robots to learn fast new tasks that do not have a concrete description, goal, or reward function \cite{ravichandar2020recent}. Arguably, despite significant progress in learning to manipulate \cite{lillicrap_continuous_2016}, the acquired skills do not generalize well across different tasks, and domains \cite{kroemer2019review}. Manipulation tasks are described by various motions related to multiple objects (i.e., contexts) \cite{kupcsik_data-efficient_2013}. Robotics research has investigated different solutions to robot manipulation relative to different specifications of the problem. Notably, motion planning methods are designated for solving problems with access to an accurate dynamics model and precise goal specification \cite{vahrenkamp2010integrated,berenson2009manipulation}. When the model is unknown or imperfect, but there is a partial task description through a reward function, reinforcement learning (RL) methods are the most suitable for acquiring complex skills \cite{haarnoja2018soft}. Imitation learning (IL) has been extensively used when neither a perfect model nor a good task description is available \cite{ravichandar2020recent}.

%Current approaches provide manipulation policies that are restricted either to a given task or a given object, and, subsequently, they do not generalize well. Indeed, even when executing the same task, e.g., object picking-and-placing, the performance is subject to different contextual parameters, e.g., the object's position and orientation, the sensory feedback, and the goal position \cite{kupcsik_data-efficient_2013}.

\begin{figure}
	\centering
	\vspace{0.8em}
	\includegraphics[height=4cm]{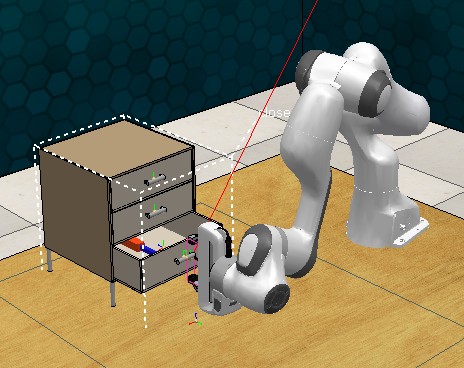}
	\includegraphics[height=4cm]{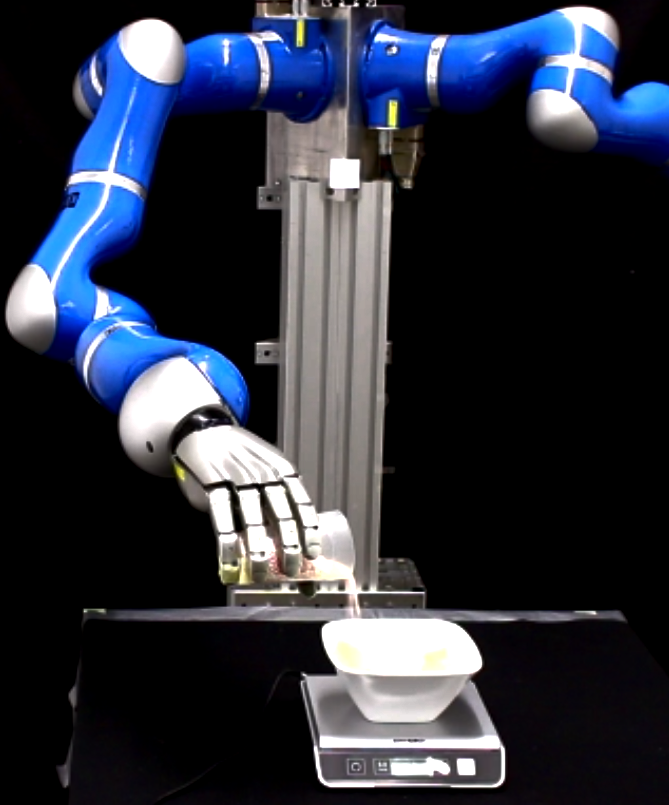}
	\vspace{-0.15cm}
	\caption{\footnotesize \textbf{Left:} Simulated drawer-closing task, where the robot should close an open drawer. \textbf{Right:} Physical robot platform performing a pouring task. In this task the robot has to learn to pour the designated amount of liquid into the bowl.}\label{fig:intro}
		\vspace{-0.7cm}
\end{figure}

IL methods employ sets of expert demonstrations on the task to be considered \cite{rueckert_extracting_2015, colome_dimensionality_2018}. Usually, these demonstrations represent a set of robotic joint configurations at specific time intervals. The most prominent methods in robot learning maps the set of demonstrated motions to a parametric model of the movements \cite{calinon2007learning, ijspeert2013dynamical}. Movement Primitives (MP) represent a broad family of dynamical system descriptors used to parameterize robot movements \cite{ijspeert2013dynamical}. Dynamic MPs (DMPs) have been thoroughly explored as a robot control policy, as they are stable and robust, however, they lack generalization properties \cite{ude2010task}. Due to their modularity, MPs emerged as a generic framework for IL, as they are sample efficient, they can be used with non-expert demonstrations, and provide safe learning \cite{ravichandar2020recent}. The framework of Probabilistic Movement Primitives (ProMPs) \cite{paraschos_probabilistic_2013, paraschos_using_2018} enjoys favorable properties, like time modulation and conditioning to different contexts, which makes it a well-suited tool for imitation and improvement of robotic movements. Contexts are typically considered as vectors, describing, for example, a goal position, the position of the object to be picked, etc. A promsing approach towards generalization of skills is the improvement of the learned policies through RL. Hence, researchers have opted for combining policy optimization with the learned MPs to generalize over different tasks, and related contexts \cite{kober2011reinforcement,stulp2011hierarchical,cheng2018fast,kang2018policy}. 

However, the usual high-dimensional representation of robotic movements complicates the application of RL techniques since learning on the real robot requires much more sample-efficiency than learning in simulation \cite{lillicrap_continuous_2016}. To overcome this difficulty, researchers have proposed the application of dimensionality reduction (DR) techniques in movement primitives (MPs) \cite{bitzer2010using,chen_efficient_2015,chen_dynamic_2016,colome_dimensionality_2014,colome_dimensionality_2014-1,colome_dimensionality_2018}. The resulting latent representation can help us decode the inter-dependencies between movements and task-contexts, allowing us to use contextual RL \cite{kupcsik_data-efficient_2013, abdolmaleki2015contextual, klink2020self} to optimize over the task-related parameters. 

A probabilistic DR technique of ProMPs in the joint space was presented in \cite{colome_dimensionality_2014}, using expectation-maximization (EM) over the context variables, for representing the movements in a linear low-dimensional subspace  of the full configuration space. Then, the policy is further optimized via relative entropy policy search (REPS) \cite{peters2010relative}. In \cite{colome_dimensionality_2014-1}, the authors introduced a DR technique of the exploration parameters of DMPs when using a path integral policy improvement algorithm \cite{theodorou2010reinforcement}. In \cite{rueckert_extracting_2015} a low  dimensional latent variable model for ProMPs is extracted using fully Bayesian hierarchical models, which is used only for imitation learning. 
Autoencoded DMPs are proposed in \cite{chen_efficient_2015}, which uses deep autoencoders to find a latent representation
of the movement from the robot's task space, towards generalizing the performance of the DMPs. In \cite{chen_dynamic_2016} the authors proposed the use of a time-dependent variational autoencoder to address the generalization challenges. Auto-encoders with ProMPs for efficient human motion prediction were proposed in  \cite{dermy2018prediction}. In \cite{colome_dimensionality_2018-1}, a reduction of the parametric space was proposed, as more appropriate for learning MPs. In \cite{delgado2020sample} the authors use parametric DR to learn a mapping from the MPs latent space to a reward for policy improvement. 

% Often the policy is warm-started by the demonstrated samples, which is further improved through policy iteration \cite{cheng2018fast}, or the demonstrated and the on-policy samples are mixed \cite{balakrishna2020policy}. However, the importance of generalizing skills led to the introduction of contextual RL methods \cite{klink2020self}. Indeed, contextual policy search has proven to be a data-efficient method for generalizing robot skills over contexts that represent either objectives of
%the robot or physical properties of the environment \cite{kupcsik_data-efficient_2013,abdolmaleki2015contextual}. The most relevant line of research to ours is the one from Colom\'{e} et al. \cite{colome_dimensionality_2018-1}, which uses contextual REPS for optimizing the policy acquired over demonstrations. 

This paper introduces a novel RL algorithm for optimizing over MPs, named LAtent-Movements Policy Optimization (LAMPO).
This algorithm's primary focus is to gain sample efficiency by 1. performing the policy updates in a reduced latent space, and 2. using off-policy gradient estimations, which re-use samples collected from previous learning episodes. In robot learning, the reduced number of samples and the use of contextual policies are important for acquiring manipulation skills through demonstrations and, subsequently, improving them so that they generalize across various contexts.
 For learning the latent representation, we use a Mixture of Probabilistic Principal Component Analysis (MPPCA) \cite{tipping_mixtures_1999}. This method, differently from other DR techniques, is fully probabilistic, allowing us to perform conditioning. MPPCA can be seen as a Gaussian mixture model (GMM), enabling us to represent multimodalities and non-linear dependencies in the demonstrated data, while it can perform de-noising, as it can extract isotropic noise contained in the data. Our method does not use demonstrations as means of initialization. We pre-train a structured latent space of the MPs through MPPCA, which is essential for exploring and optimizing through RL.

For the off-policy estimation we use self-normalized importance sampling \cite{owen_monte_2013}. Differently from other approaches \cite{shelton_policy_2001,peshkin_learning_2002}, we perform a full-gradient estimation, also considering the normalization factor, which further lowers the variance, similarly to the baseline subtraction method \cite{deisenroth2013survey, jie2010connection}. Off-policy evaluation and improvement are core problem in RL. The off-policy estimation is in fact hard to obtain, due to distribution mismatch \cite{thomas2016data,shelton_policy_2001}. Recent model-free deep RL methods employ off-policy estimation \cite{haarnoja2018soft}; though, such methods require a vast amount of interactions of the agents with the environment. Notably, our method combines off-policy estimation together with trust-region regularization that contains the estimation variance \cite{peters_relative_2010,schulman_trust_2015, schulman2017proximal}. We apply a forward Kullback-Leibler (KL) divergence bound between the subsequent policies, which, jointly to a KL-regularization of the context distribution, provides a robust policy optimization. We evaluated our algorithm both in simulated environments provided by RLBench \cite{james_rlbench_2020} and on a real manipulator robot. The experimental analyses show that LAMPO outperforms the state of the art techniques in sample efficiency on challenging high-dimensional robotic manipulation tasks.

%\sam{I think we should also clarify one step. Usually one initialize the MP with the expert data and then to modify the parameters via RL. Instead, we want to initialize a proper latent space of the MP parametrization, to find a good space where to optimize (or explore) with RL. I know that the concept is difficult to say, but I think this distinguish a lot our work from many others.}

%\section{Related Work}
%\input{sections/related_work.tex}

\section{Problem Statement}
\begin{figure}[t]
	\begin{subfigure}[t]{0.95\columnwidth}
		\centering
		\input{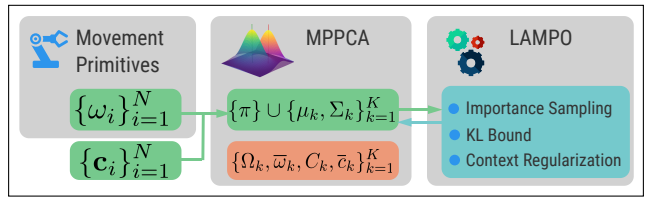}
	\end{subfigure}
	\vspace{-0.15cm}
	\caption{\footnotesize A graphical description of our algorithm. The movement's parameters and the context are projected to a latent space using MPPCA. The proposed off-policy method optimizes over the latent space of movements leading to policy improvement.} 
	\label{fig:graphical_models}
		\vspace{-0.7cm}
\end{figure}     
We consider contextual problems where the robot must adapt its behaviour based on a context $\cvec$. The context can be seen either as a description of the current state of the task (e.g., position of an object), or as a goal (e.g., a desired quantity that the robot should pour in a bowl) and is described by a $d_c$-dimensional vector (i.e., $\cvec \in \mathbb{R}^{d_c}$). At each episode, the robot observes a new context distributed according to $q(\cvec)$. In our work, we consider robotic movements parametrized by a vector of parameters $\omegavec \in \mathbb{R}^m$. Since the movements depend on the context $\cvec$, the behavior of the human demonstrator can be described as a stochastic mapping $q(\omegavec | \cvec)$ between the movements' parameters $\omegavec$ and the context $\cvec$. A user-defined reward $R(\omegavec, \cvec)$ indicates how well a particular movement parametrized  by $\omegavec$ performed according to the context $\cvec$. Our objective is to find the optimal policy which maps movements and contexts
%, and maximizes the average reward, according to the following optimization problem \color{black}
%	}
%\sout{
%We consider a context distribution $q(\cvec)$ with $\cvec \in \mathbb{R}^{d_c}$, a known mapping $\mathcal{M}\mathcal{P}(\omegavec)$ from a parametric space $\mathbb{R}^{m}$ to a robotic trajectory, an unknown probabilistic mapping $q(\omegavec | \cvec)$ from the context $\cvec$ to a parametric movement $\omegavec$, which represents the behavior of the human demonstrator; and a user-defined reward function $R(\omegavec, \cvec)$.
%The goal is to find the policy $p_\theta(\omegavec | \cvec)$ that maximizes the average reward}
%\small
\begin{align}
	\theta^* = \argmax_{\theta} \int R(\omegavec, \cvec)p_\theta(\omegavec | \cvec)q(\cvec)\de \cvec \de \omegavec \label{eq:objective-1}.
\end{align}       
\normalsize

% \geo{I think I removed this already. This is not a part of the problem. This a solution to the above problem. Please move it somewhere that is appropriate.}\sam{Sorry, I didn't see you deleted it. But where would you put it? I tried to do something similar to most of RL papers, where they put in the problem statement the objective... do you think is better to keep it only as a text? We can mention here that our objective is to maximize the reward and present this equation in the beginning of RL. so it is closer to where we elaborate on this :-) .}
% Usually, as already mentioned, we learn an initial policy via imitation learning, where we aim to find $\theta_0$ for which $p_{\theta_0}(\omegavec | \cvec)$ best approximates $q(\omegavec | \cvec)$. This is usually done by maximizing 
% \begin{align}
% 	\hat{\mathcal{L}} = \sum_i \log p_{\theta_0}(\omegavec_i,  \cvec_i) \label{eq:likelihood}
% \end{align}
% where $\omegavec_i,  \cvec_i \sim q(\omegavec_i | \cvec_i)q(\cvec_i)$, and then we can use conditioning to find $p_{\theta_0}(\omegavec | \cvec)$.
% Once we obtain a good initial policy $p_{\theta_0}$, we aim to use reinforcement learning to improve it. This is done by using a generative model of $p_(\omegavec | \cvec)$, and at every iteration $T$ we collect a set $\{R_{i}, \omegavec_{i}, \cvec_{i}\}_{i=n(T-1)}^{nT}$ of $n$ episodes using the set of parameter $\theta_{T-0}$ and deriving a better set of parameters $\theta_T$. In our framework, as we will see, we will replace the high-dimensional $\omegavec_{i}$ with a better suited low-dimensional variable.

\section{Latent Movements Policy Optimization}
Finding the optimal policy of the objective \eqref{eq:objective-1}, is non-trivial. In our setting, we acquire an initial set of $N$ demonstrations, from which we learn an initialization of the parameters of the ProMPs.
We start with IL, where we aim to find an initial policy parameter set $\theta_0$ 
%(where the subscript $0$ indicates that this set of parameters is the first of a series of improvement carried out with RL) 
for which the initial policy $p_{\theta_0}(\omegavec | \cvec)$ approximates the demonstrator's behavior $q(\omegavec | \cvec)$. By maximizing 
\begin{align}
	\theta_0 = \argmax_{\theta}  \sum_i \log p_{\theta_0}(\omegavec_i,  \cvec_i), \label{eq:likelihood}
\end{align}
where $\omegavec_i,  \cvec_i \sim q(\omegavec_i | \cvec_i)q(\cvec_i)$, we can then  find the policy $p_{\theta_0}(\omegavec | \cvec)$ by conditioning.
Subsequently, we aim to use RL for policy improvement. This requires the use of a generative model of $p(\omegavec | \cvec)$ for generating trajectories over context. At every iteration $T$, we collect a set of rewards $\{R_{i}, \omegavec_{i}, \cvec_{i}\}_{i=(T-1)n}^{Tn}$ for $n$ number of episodes using the $T^{th}$ set of parameters $\theta_T$ and optimize objective \eqref{eq:objective-1} to acquire a better set of parameters $\theta_{T}$ (Alg.~\ref{alg:lampo}). In our framework, we replace the high-dimensional $\omegavec_{i}$ with a better suited low-dimensional latent representation.
Fig. \ref{fig:graphical_models} is a graphical description of the proposed algorithm, which we detail in the following.

\subsection{Imitation Learning with Dimensionality Reduction}
%\sam{If you agree, Georgia, we can skip all of these details and referenced to Parachos, and just underline that MP is a parametrization of the movement which allows us to define a movement based on the parameter vector $\omegavec$.}
%\sam{SIMPLER VERSION: 
Consider the specific parametrization of the MP in \cite{paraschos_probabilistic_2013}, which allows to obtain a specific smooth trajectory $\bm{\tau}$  depending on a parameter vector $\omegavec$. Given a specific trajectory $\bm{\tau}$, is possible to infer the most likely set of parameters $\bm{\omega}$ that generated such trajectory (via Ridge regression). The ProMPs framework introduces the density estimate  of the parameters $p(\omegavec)$ given a set of demonstrations $\{\bm{\tau}\}$. Furthermore, using probabilistic conditioning, one can provide a set of demonstrations that depends on a context $\{\bm{\tau_i, \cvec_i}\}$ and infer what are the most likely parameters given a new unsen context $\cvec$, i.e., $p(\omegavec | \cvec)$.
To overcome the high dimensionality of ProMPs, we introduce a mixture of probabilistic movements to find a latent space to represent jointly the movement parameters and the context, % {\color{red}Hence, we introduce the following graphical model based on MPPCA: }
\begin{equation}
	\begin{array}{cclr}
	k &\!\sim\! &  Cat(k | \pivec) & (k \in \{1, \dots K\})  \nonumber \\
	\zvec &\!\sim\!  & \mathcal{N}(\zvec | \muvec_k, \Sigmavec_k) &  (\zvec \in \mathbb{R}^{d_z}) \nonumber \\
	\epsvec_{\omega}, \epsvec_{c} & \!\sim\! & \mathcal{N}(\epsvec_\omega, \epsvec_c|\mathbf{0},  \mathbf{I})  &  \nonumber\\
	\omegavec &\!=\! & \Omegavec_k \zvec + \overline{\omegavec}_k +\sigma_k^2 \epsvec_{\omega}  &  (\bm{\omega} \in \mathbb{R}^m)\nonumber \\
	\cvec & \!=\! & \Cvec_k \zvec + \overline{\cvec}_k +\sigma_k^2 \epsvec_{c}  & (\cvec \in \mathbb{R}^{d_c})\nonumber
	\end{array}
\end{equation}

\noindent where $k$ is a categorical latent variable representing the selection of a particular Gaussian, its parameter $\pivec$ denotes the probabilities describing this selection; $\zvec$ is a Gaussian latent variable representing a specific movement-context pair in each cluster. The isolation of the isotropic noises $\epsvec_{\omega}, \epsvec_{c}$ helps to compress only the useful information, and act as a denoiser in the generative model.

Groping the two last equations together 
\begin{equation}
	\left[\begin{matrix}
	\omegavec \\
	\cvec 
	\end{matrix}\right] = \left[\begin{matrix}
	\Omegavec_k \\
	\Cvec_k 
	\end{matrix}\right]\zvec + \left[\begin{matrix}
	\overline{\omegavec}_k \\
	\overline{\cvec}_k 
	\end{matrix}\right] + \sigma_k^2 \left[\begin{matrix}
	\epsvec_{\omega} \\
	\epsvec_{c}
	\end{matrix}\right],\label{eq:model}
\end{equation}
we notice the equivalence with the mixture of principal component analyzers.

The parameters $\{\pivec, \Omegavec_k, \Cvec_k, \overline{\omegavec}_k, \overline{\cvec}_k, \sigma_k^2\}$ can be inferred via EM  \cite{tipping_mixtures_1999}, assuming, without loss of generality, that $\muvec_k=\bm{0}, \Sigmavec_k=\mathbf{I}$.
Once the maximum-likelihood parameters are obtained, %we can generate the movement parameters $\omegavec$ given a context $\cvec$. 
to generate new movements conditioned on a context $\cvec$, we first sample the latent variables $\zvec, k$ conditioned on $\cvec$, and then we establish the relation $\omegavec = \Omegavec_k \zvec + \overline{\omegavec}_k$ removing the isotropic noise on the movement (as it is an unnecessary perturbation).

The generative model of $p(\zvec, k | \cvec)$ is a GMM, $p(\zvec, k | \cvec) = p(\zvec | k, \cvec)p(k | \cvec)$,
where the conditional responsibility is
\begin{align}
	p(k | \cvec) = \frac{p(\cvec| k)\pi_k}{\sum_k p(\cvec|k)\pi_k}
\end{align}
and $p(\cvec|k)$ is obtained marginalizing $\zvec$ from $p(\cvec|k, \zvec)$ in \eqref{eq:model} and 
\begin{align}
p(\zvec | k, \cvec) & = &\!\mathcal{N}\left(\zvec | \Bvec_k \left(\sigma_k^2\Cvec^\intercal_k (\cvec - \overline{\cvec}_k) + \Sigmavec_k\muvec_k\right), \Bvec_k \right)
\end{align}
with $\Bvec_k = \left(\Sigmavec_k + \sigma^{-2}_k\Cvec_k^\intercal\Cvec_k\right)^{-1}$.
In the generative model, we assume no isotropic noise on the movement's parameter
\begin{align}\label{eq:generative}
	\omegavec = \Omega_k\zvec +\overline{\omegavec}_k \quad \text{with} \quad \zvec \sim p(\zvec | k, \cvec), \quad k \sim p(k | \cvec).
\end{align}
The described generative model captures non-linear dependencies between the context and the robotic movement, both defining a convenient latent representation and maintaining mathematical tractability.

\subsection{Off-Policy Reinforcement Learning}
% In the previous section we {\color{red} showed how to train} a generative model {\color{red} based on} a mixture of probabilistic principal component analyzers. 
The variables $\{\Cvec_k, \overline{\cvec}_k, \Omegavec_k, \overline{\omegavec}_k, \sigma_k\}_{k=1}^{K}$ describe how to project, for each component $k$, a variable $\zvec$ to a movement-context pair. To maintain a plausible representation of the movement, this projection should remain fixed, while we optimize the distributions of $k$ and $\zvec$ that represent how the movements and the context are distributed in the latent space. The variables $k$ and $\zvec$ are described by $\{\pivec\} \cup \{\muvec_k, \Sigmavec_k\}_{k=1}^K$.
%\geo{How is this different from the previous section? Please explain what these variables now represent.}
Since the model's variables $\{\Omegavec_k, \overline{\omegavec}_k, \Cvec_k, \overline{\cvec}_k\}_{k=1}^K$ are now fixed, they can be considered a part of the reward signal, changing the initial problem \eqref{eq:objective-1} into 
\begin{align}
	J(\theta) & = \int R(\omegavec, \cvec)p_\theta(\omegavec | \cvec)q(\cvec)\de \omegavec\de \cvec \nonumber \\
	& = \int \sum_{k=1}^m R(\Omegavec_k \zvec + \overline{\omegavec}_k, \cvec)p_\theta(\zvec, k | \cvec)q(\cvec) \de \cvec. \label{eq:objective}
\end{align}
The above formulation provides a desirable latent representation of the policy, as the latter is depended on the latent parameters $\zvec, k$ conditioned on the underlying context $\cvec$.
%To obtain an efficient and reliable policy optimization process, we consider the following steps. (1) 
LAMPO interleaves policy-optimization and data-gathering.
At each iteration $T$ we consider all samples collected from all past policies $\theta_1, \dots, \theta_{T-1}$. For each new set of parameters $\theta_T$, we obtain a new conditional model $p_{\theta_T}(\omega | \cvec)$, and a new context distribution $p_{\theta_T}(\cvec)$ that allows to obtain new samples. However, when $p_{\theta_T}(\cvec)$ diverges from $q(\cvec)$, the computation of $p_{\theta_T}(\omega | \cvec)$ can suffer from numerical instabilities. To prevent this issue, we use a KL-regularization to keep the two distributions close enough. Furthermore, to prevent premature convergence of the policy to a local optimum, we use a KL-constraint between the previous and the current policy distribution i.e., $KL(p_{\theta}(\cdot|\cvec)\| p_{\theta_T}(\cdot||\cvec))$. 
\subsubsection{Off-Policy Estimate}
To reuse all the past experience, we propose to use Self-Normalized Importance Sampling (SNIS) \cite{owen_monte_2013}. SNIS estimation has usually lower variance'than pure importance sampling, at the price of a small bias. Its usage in the context of RL is well established \cite{shelton_policy_2001,peshkin_learning_2002}.

At the iteration $T+1$, the dataset is composed of samples generated by $T$ different policies.% parametrized with $\theta_1, \dots, \theta_T$.
% \sout{During the generation of the dataset, we keep information of which policy has generated which sample. 
%%Formally, let us consider the dataset $\{R_{i}, \zvec_{i}, k_{i}, j_i\}_{i=1}^{Tn}$ composed by $Tn$ different samples, each one of those generated by sampling uniformly an index $j_i$ from $\{1, \dots, T\}$, $\cvec \sim q(\cvec)$ and  subsequently $\zvec_i, k_i \sim p_{\theta_{j_i}}(\zvec_i, k_i|\cvec_i)$. 
%Now, suppose you want to correct the distribution of a sample $\{\zvec_i, k_i, \cvec_i\}$ generated by the policy $\theta_\zeta$. Formally, $\theta_\zeta$ has been chosen uniformly from $\{\theta_1, \dots, \theta_T\}$, with $\zeta$ being just a selction index, then $\cvec_i \sim p(\zvec_i, k_i |\cvec_i, \zeta) = p_{\theta_\zeta}(\zvec_i, k_i |\cvec_i)$.
%\sam{I would suppress $\rho_i$ and just describe it.}
%With the above assumptions, we can compute the importance sampling ratio}
{We compute the importance ratio $\rho_i$ for each pair of samples $\zvec_i, k_i$ by using the current policy $p_\theta(\zvec_i, k_i |  \cvec_i)$ at the numerator and the mixture of the past policies at the denominator. The SNIS estimate results to be}
%\begin{align}
%	\rho_{i} & = \frac{p_\theta(\zvec_{i}, k_{i},\cvec_i, \zeta)}{p(\zvec_{i}, k_{i},\cvec, \zeta)} = \frac{p_\theta(\zvec_{i}, k_{i}|\cvec_i)p(\zeta)q(\cvec_i)}{p(\zvec_{i},
%			k_{i}|\cvec_i , \zeta)p(\zeta)q(\cvec_i)} \nonumber \\
%	 & = \frac{p_\theta(\zvec_{i}, k_{i}|\cvec_i)p(\zeta)q(\cvec_i)}{p_{\theta_{\zeta}}(\zvec_{i},
%	  k_{i}|\cvec_i)p(\zeta)q(\cvec_i)} = \frac{p_\theta(\zvec_{i}, k_{i}|\cvec_i)}{p_{\theta_{\zeta}}(\zvec_{i}, k_{i}|\cvec_i)} \nonumber 
%\end{align}
%where $p(\zeta)=1/T$, then (\ref{eq:objective}) yields to\footnote{The dependencies of $\nu$ and $\rho_i$ from $\theta$ are omitted for simplicity, however this dependency should be considered when computing the gradient.}
\begin{align}
	J(\theta) \approx \hat{J}(\theta) =  \sum_{i=1}^{nT} \frac{\rho_{i}}{\nu} R_{i} \mbox{ with }\nu = \sum_{i=1}^{nT}\rho_{i}. \label{eq:approxobj}
\end{align}

\subsubsection{Context Regularization}
%We introduce a regularization over the approximation of the contextual distribution. 
To compute $p(\zvec, k | \cvec)$, we need the ratio $p(\zvec, k, \cvec)/p(\cvec)$. %
Let us rewrite %For simplicity of notation we return back to the objective
 \eqref{eq:objective} as %, which for this step can be rewritten as
\begin{align}
	\int \sum_{k=1}^K R(\Omegavec_k \zvec + \overline{\omegavec}_k, \cvec)\frac{p_\theta(\zvec, k, \cvec)}{p_\theta(\cvec)}q(\cvec) \de \cvec.
\end{align} 

%After the IL phase, we acquire an approximation of the context distribution $p_\theta(\cvec) \approx q(\cvec)$. However, during the optimization of $\theta$, the distribution $p_\theta$ eventually changes, thus, affecting the contextual distribution approximation.
During the policy optimization, $p_\theta(\cvec)$ can diverge from $q(\cvec)$ causing high variance of the estimate.
To avoid this issue, we encourage $p_\theta(\cvec)$ to stay close to $p_{\theta_0}(\cvec)$ using the KL-divergence, where $\theta_0$ are the initial policy's parameters acquired with IL.  
Since $p_\theta(\cvec)$ is a mixture of Gaussians, we cannot compute in closed form the KL between $p_\theta(\cvec)$ and $p_{\theta_0}(\cvec)$. To overcome this issue, we compute the KL between $p_\theta(\cvec, k)$ and $p_{\theta_0}(\cvec, k)$, which is an upper-bound of the previous term
\begin{align}
& \int \sum_kp_\theta(\cvec, k) \log\frac{p_\theta(\cvec, k)}{p_{\theta_0}(\cvec, k)}\de \cvec \nonumber \\
=& \int \underbrace{p(\cvec)\sum_k p_\theta(k|\cvec)\!\log\! \frac{p_{\theta}(k|\cvec)}{p_{\theta_0}(k|\cvec)}}_{\text{Always non-negative}}\!\de \cvec +\! \int\!  p_\theta (\cvec)\! \log\! \frac{p_\theta (\cvec)}{p_{\theta_0} (\cvec)}\!\de \cvec. \nonumber 
\end{align}
The analytical expression of the KL-divergence between $p_\theta(\cvec, k)$ and $p_{\theta_0}(\cvec, k)$ takes the form of 
\small
\begin{align}
	&\eta_\theta =  KL\left(p_{\theta}(\cvec, k) \| p_{\theta_0}(\cvec, k) \right)\nonumber  \\
%	\\ & = \int \sum_k p_\theta(\cvec, k)\log \frac{p_{\theta}(\cvec, k)}{p_{\theta_0}(\cvec, k)}\de \cvec \nonumber \\ 
%	& = p_\theta(k|\cvec) \int \sum_k p_\theta(\cvec| k)\bigg( \log p_\theta(\cvec| k)  \nonumber  \\
% & + \log p_\theta(k) -\log p_{\theta_0}(\cvec| k) - \log p_{\theta_0}(k)\bigg) \de \cvec \nonumber  \\ 
	& = \sum_k p_{\theta}(k)\bigg(\mathcal{H}\left(p_{\theta}(\cvec|k),p_{\theta_0}(\cvec|k) \right)  - \mathcal{H}\left(p_{\theta}(\cvec|k),p_{\theta_0}(\cvec|k) \right) \bigg)\nonumber \\ &= + \mathcal{H}\left(p_\theta(k), p_{\theta_0}(k)\right)   - \mathcal{H}\left(p_\theta(k), p_{\theta}(k)\right) \label{eq:regularization}
\end{align}
\normalsize
where $\mathcal{H}$ is the entropy, and since $p_{\theta}(\cvec)$ and $p_{\theta_0}(\cvec)$ are Gaussian distributions, and $p_{\theta}(k)$ and $p_{\theta_0}(k)$ are categorical distributions, \eqref{eq:regularization} is computable in closed form.
%\begin{align}
%	\mathcal{H}\left(p_{\theta}(\cvec|k),p_{\theta_0}(\cvec|k) \right)
%	CH_k & = -\int p_{\theta}(\cvec,  k)\log p_{\theta_0}(\cvec | k)\de \cvec \nonumber \\
%	H_k & = -\int p_{\theta}(\cvec | k)\log p_{\theta}(\cvec | k)\de \cvec \nonumber \\
%	CH & = - \sum_k p_{\theta}(k)\log p_{\theta_0}(k),	H  = - \sum_k p_{\theta}(k)\log p_{\theta}(k).\nonumber
%\end{align} 
%Since $p_{\theta}(\cvec)$ and $p_{\theta_0}(\cvec)$ are Gaussian distributions, and $p_{\theta}(k)$ and $p_{\theta_0}(k)$ are categorical distributions, $CH_k, H_k, CH$ and $H$ can be computed in closed-form.
%The cross-entropy between two multivariate normal distribution $p_A$ and $p_B$ with parameters respectively $\muvec_A, \Sigmavec_A$ and $\muvec_B, \Sigmavec_B$ is
%\begin{align}
%	 & H(p_A,p_B) 
%	= \frac{1}{2}\bigg(d\log(2\pi) + \mathrm{Tr}\left(\Sigmavec_B^{-1}(\muvec_A\muvec_A^\intercal + \Sigmavec_A)\right)\nonumber \\
%	&  + \log|\Sigmavec_B|  - \muvec_B^\intercal\Sigmavec_B^{-1}\muvec_A
%	- \muvec_A^\intercal\Sigmavec_B^{-1}\muvec_B + \muvec_B^\intercal\Sigmavec_B^{-1}\muvec_B  \bigg).
%\end{align}
%It is possible to substitute  $\muvec_A, \muvec_B, \Sigmavec_A, \Sigmavec_B$ with the parameters of $p_\theta(\cvec|k)$ and $p_{\theta_0}(\cvec | k)$ to obtain the values of $H_k$ in closed form.
Using the divergence in \eqref{eq:regularization} as a regularization term in our objective \eqref{eq:approxobj} stabilizes the optimization process.

\subsubsection{Trust Region}
To obtain a robust optimization process, it is a common practice to introduce a KL-constraint (or regularization) between two subsequent policies during the optimization \cite{peters2010relative,schulman_trust_2015}. These methods avoid the premature convergence to a local optimum and they guarantee a smooth optimization process.
Usually, the KL-constraint is approximated via the samples generated by the previous policy \cite{peters_relative_2010}.
%In our work, however, the samples are originated from all previous policies, making the above approach not applicable. 
in our case the KL between the current and the previous policy defined on the latent variables $k$ and $\zvec$ have closed form, as shown in the previous subsection.
Hence, the KL divergence between the optimized policy and the policy at iteration $T-1$, can be computed as
\begin{align}
	g_\theta(\cvec) =  KL\left(p_{\theta_{T-1}}(\cdot | \cvec) \| p_{\theta}(\cdot | \cvec) \right). \label{eq:constraint}
\end{align}

\subsection{Off-policy improvement}
We have introduces so far the off-policy objective, the context and the the trust region regularization. We can formulate the gradient estimation as a constrained optimization problem. To ensure the parameters consistency, we encode the categorical distribution as $\pivec = e^{\theta_i}/\sum_{i=1}^Ke^{\theta_i}$, and the covariances $\Sigmavec_k$ as positive-definite diagonal matrices.
By combining \eqref{eq:approxobj} \eqref{eq:regularization} and \eqref{eq:constraint}, we obtain
\begin{align}
	 \max_{\theta} \hat{J}(\theta) - \gamma \eta_\theta 
	 \text{ s.t. } (nT)^{-1} \sum_{i=1}^{Tn} g_\theta(\cvec_i) \leq \chi \label{eq:optproblem}
\end{align}
where $\chi$ is the upper-bound of the forward KL, and $\gamma$ is the regularization constant which controls the impact of the regularization on the objective. 
All the entities composing problem \eqref{eq:optproblem} have analytical gradient, allowing us to use Sequential Least SQuare Program (SLSQP) optimization. 

The gradient w.r.t. the approximated objective \eqref{eq:approxobj} is
\small{
\begin{align}
 	\nabla_{\theta}\hat{J}(\theta) &  = \nu^{-1}\sum_{i=1}^{nT} \left( \nabla_\theta \rho_{i} - \nu^{-1}\rho_{i} \nabla_\theta\nu  \right)    R_{i} \nonumber \\
 	& = \sum_{i}\frac{\rho_{i}}{\nu}\nabla_\theta \log p_\theta(\zvec_{i}, k_{i}|\cvec_{i}) \left(R_{i} - \hat{J}(\theta)\right),
\end{align}}
\normalsize
which turns out to be equal to the SNIS of the classic gradient with baseline subtraction, yielding lower variance in the gradient estimation \cite{williams_simple_1992, sutton_policy_2000, deisenroth2013survey}.
We could compute $\nabla_\theta \log p_\theta(\zvec_{i}, k_{i}|\cvec_{i})$ by taking the gradient of \eqref{eq:generative}, but due to the matrix inversion in $\Bvec_k$, the computation presents numerical instability. We propose, instead, the derivation of the gradient from 
\begin{align}
	\log p_\theta(\zvec, k |\cvec) & = \log p_\theta(\cvec | \zvec, k ) +  \log p_\theta(\zvec | k) + \log \pi_k \nonumber \\
	& - \log \sum_j \pi_j \underbrace{\int p_\theta(\cvec | \zvec, j ) p_\theta(\zvec | j)\de \zvec}_{p_\theta(\cvec | j)}
\end{align}
which is solvable in closed form, where $p_\theta(\cvec | j) = \mathcal{N}\left(\cvec|\Cvec_j \muvec_j + \overline{\cvec}_j, \sigma_j^2\mathbf{I} + \Cvec_j\Sigmavec_j\Cvec_j^{-1}\right)$,
does not require the inversion of $\Bvec_k$.
The gradient can be computed via automatic differentiation tools, like \texttt{PyTorch} or \texttt{Tensorflow}. An overall description of LAMPO is presented in Alg. \ref{alg:lampo}.

\section{Experimental Analysis}
\subsection{Experimental setup}\label{sec:details}
\begin{algorithm}[t] \vspace{-0.1cm}
	\caption{LAtent Movement Policy Optimization}
	\label{alg:lampo}
	\footnotesize
	\begin{algorithmic}[1]
		\State \textbf{input:} Dataset $\{\bm{\tau}_i, \cvec_i\}_{i=1}^{n_d}$ of trajectories and contexts.
		\State \textbf{for each} trajectory $\bm{\tau}_i$ compute the parameter vector $\bm{\omega_i}$.
		\State EM to find the MPPCA's parameters $\pivec, \Omegavec_k, \overline{\omegavec_k}, \Cvec_k, \overline{\cvec}_k$ \cite{tipping_mixtures_1999}\\
		assuming $\muvec_k=\bm{0}$, $\Sigmavec_k=\mathbb{I}$.
		\State Set $\theta_0 \equiv \{\pivec\} \cup \{\muvec_k, \Sigmavec_k\}_{k=1}^K$.
		\For{$T \in \{1 \cdots N_T\}$}
		\For{$i \in \{n(T-1) \dots nT\}$}
		\State Observe context $\cvec_i$
		\State Perform movement $\omegavec_i$ with \eqref{eq:generative}
		\State collect reward $R_i$
		\EndFor
		\State Compute a differentiable model of $\hat{J}(\theta)$, $\eta_\theta$ and $g(\theta)$ using \eqref{eq:approxobj}, \eqref{eq:regularization} and \eqref{eq:constraint}.
		\State Solve \eqref{eq:optproblem} for $\{\pivec\} \cup \{\muvec_k, \Sigmavec_k\}_{k=1^K}$ using SLSQP.
		\State Update $\theta_T \equiv \{\pivec\} \cup \{\muvec_k, \Sigmavec_k\}_{k=1}^K$.
		\EndFor
	\end{algorithmic}
\end{algorithm}	 
In the following, we analyze the performance of LAMPO. We study the effect of motion multimodalities, comparing against baseline approaches, i.e. planning, deepRL and policy improvement methods. We also test whether LAMPO scales to different contexts and dimensionalities of a given problem. Finally, we train and validate our proposed method on a real robotic platform.
%We have conducted a series of experiments both in virtual environments and on a real manipulator robot.
%In what follows, we describe in detail the experimental setup of these evaluations. \\

\noindent \textbf{{Virtual environments.}}\\
\noindent\texttt{2d-reacher:} toy simulated environment where a robotic manipulator is composed of two revolute joints, and two links of length one.
The task consists in reaching four different target areas, as depicted in Fig.~\ref{fig:reacher2d}-left. The user can select the number of clusters from which one can generate the respective goals, to acquire diverse and of different complexity movements in the dataset. The demonstrations are provided using inverse kinematics (IK). A variation of the previous task includes also an obstacle (Fig.\ref{fig:obstacle}). The purpose is to compare LAMPO performance against RRT*. The demonstrations are also provided using RRT*.
% The demonstrations are provided via inverse kinematics (IK), while for providing sub-optimal demonstrations, the goal position is perturbed with a xero mean Gaussian noise with standard deviation $\sigma^2=0.15$.
The movements are encoded by $20$ radial basis functions, for a total of $40$ parameters. \color{black} We add to this set an extra-parameter to encode the duration of the movement (we will keep a similar setting in all the experiments).
% 
% The context variable is $2$-dimensional, representing the goal position. The reward function is defined by the euclidean distance between the robot end-effector and the goal position. 
 \begin{figure}[t]
 	\vspace{-0.15cm}
 	\centering
 	\begin{subfigure}{0.32\columnwidth}
 		\input{figures/2dreacher.tex}
 	\end{subfigure}
 	\hspace{-0.3cm}
 	\begin{subfigure}{0.32\columnwidth}
 		\vspace{0.10cm}
 		\input{figures/reacher-rlbench-env.tex}
 	\end{subfigure}
 	\hspace{-0.0cm}
 	\begin{subfigure}{0.32\columnwidth}
 		\vspace{0.10cm}
 		\input{figures/reacher-darias-env.tex}
 	\end{subfigure}
 	\vspace{-0.5cm}
 	\caption{\footnotesize \textbf{Left:} Our \texttt{2d-reacher}.The goal is to reach some given point.Tthe number of clusters of the goal-positions can be modified to generate datasets with different degrees of non-linearities. \textbf{Center:} The \texttt{reach-target} from RLBench. \textbf{Right:} Our real-robot reacher. \label{fig:reacher2d}}
 	%\vspace{-0.3cm}
 	\hspace{-0.45cm}
 	\begin{tikzpicture}

\definecolor{color0}{rgb}{1.0,0.25,1.0}
\definecolor{color1}{rgb}{0.,0.25,1}

\begin{axis}[
width=0.6\columnwidth,
legend cell align={left},
legend style={fill opacity=0.8, draw opacity=1, text opacity=1, at={(0.97,0.03)}, anchor=south east, draw=white!80!black, font=\scriptsize},
tick align=outside,
tick pos=left,
title={\small\texttt{2dreacher w/ obstacle}},
title style={yshift=-7pt},
x grid style={white!69.0196078431373!black},
xmajorgrids,
xmin=-0.7, xmax=14.7,
xtick style={color=black},
y grid style={white!69.0196078431373!black},
ymajorgrids,
ymin=0.674073768203892, ymax=1.0,
ytick style={color=black},
ylabel={\scriptsize Success Rate},
ylabel style={yshift=-10pt},
xlabel={\scriptsize \# Iterations},
yticklabel style={font=\scriptsize},
xticklabel style={font=\scriptsize},
]

\addplot [line width=2pt, color0, style=dashed]
table[row sep=\\] {%
	0 0.98\\
	14 0.98\\
};
\addlegendentry{RRT*};

\addplot [semithick, color1, forget plot]
table[row sep=newline] {%
0 0.706666666666667
1 0.736666666666667
2 0.783333333333333
3 0.843333333333333
4 0.858333333333333
5 0.836666666666667
6 0.891666666666667
7 0.893333333333333
8 0.888333333333333
9 0.92
10 0.883333333333333
11 0.901666666666667
12 0.898333333333333
13 0.913333333333333
14 0.908333333333333
};
\path [draw=color0, semithick, dash pattern=on 5.55pt off 2.4pt]
(axis cs:0,0.98)
--(axis cs:14,0.98);

\path [draw=color1, semithick]
(axis cs:0,0.688641684003707)
--(axis cs:0,0.724691649329626);

\path [draw=color1, semithick]
(axis cs:1,0.712633356449387)
--(axis cs:1,0.760699976883946);

\path [draw=color1, semithick]
(axis cs:2,0.749286143858854)
--(axis cs:2,0.817380522807813);

\path [draw=color1, semithick]
(axis cs:3,0.819300023116054)
--(axis cs:3,0.867366643550613);

\path [draw=color1, semithick]
(axis cs:4,0.851323617853293)
--(axis cs:4,0.865343048813373);

\path [draw=color1, semithick]
(axis cs:5,0.820644459855147)
--(axis cs:5,0.852688873478186);

\path [draw=color1, thick]
(axis cs:6,0.872640296077987)
--(axis cs:6,0.910693037255346);

\path [draw=color1, thick]
(axis cs:7,0.893333333333333)
--(axis cs:7,0.893333333333333);

\path [draw=color1, thick]
(axis cs:8,0.875315290298974)
--(axis cs:8,0.901351376367693);

\path [draw=color1, thick]
(axis cs:9,0.91399167244568)
--(axis cs:9,0.92600832755432);

\path [draw=color1, thick]
(axis cs:10,0.863305574818934)
--(axis cs:10,0.903361091847733);

\path [draw=color1, semithick]
(axis cs:11,0.880637520226547)
--(axis cs:11,0.922695813106786);

\path [draw=color1, thick]
(axis cs:12,0.893326393704733)
--(axis cs:12,0.903340272961933);

\path [draw=color1, thick]
(axis cs:13,0.903319454076133)
--(axis cs:13,0.923347212590533);

\path [draw=color1, thick]
(axis cs:14,0.891309738596094)
--(axis cs:14,0.925356928070573);

\addplot [semithick, color1, mark=-, mark size=5, mark options={solid}, only marks, forget plot]
table[row sep=newline] {%
0 0.688641684003707
1 0.712633356449387
2 0.749286143858854
3 0.819300023116054
4 0.851323617853293
5 0.820644459855147
6 0.872640296077987
7 0.893333333333333
8 0.875315290298974
9 0.91399167244568
10 0.863305574818934
11 0.880637520226547
12 0.893326393704733
13 0.903319454076133
14 0.891309738596094
};
\addplot [semithick, color1, mark=-, mark size=5, mark options={solid}, only marks, forget plot]
table[row sep=newline]  {%
0 0.724691649329626
1 0.760699976883946
2 0.817380522807813
3 0.867366643550613
4 0.865343048813373
5 0.852688873478186
6 0.910693037255346
7 0.893333333333333
8 0.901351376367693
9 0.92600832755432
10 0.903361091847733
11 0.922695813106786
12 0.903340272961933
13 0.923347212590533
14 0.925356928070573
};
\addplot [line width=2pt, color1]
table[row sep=newline] {%
0 0.706666666666667
1 0.736666666666667
2 0.783333333333333
3 0.843333333333333
4 0.858333333333333
5 0.836666666666667
6 0.891666666666667
7 0.893333333333333
8 0.888333333333333
9 0.92
10 0.883333333333333
11 0.901666666666667
12 0.898333333333333
13 0.913333333333333
14 0.908333333333333
};
\addlegendentry{LAMPO}
\end{axis}
\end{tikzpicture}
 	\begin{subfigure}{0.39\columnwidth}
 		\vspace{-5.0cm}
 		\input{figures/obstacle-env.tex}
 	\end{subfigure}
 	\vspace{-0.3cm}
 	\caption{\label{fig:obstacle}\footnotesize \textbf{Left} learning curve of LAMPO in the \texttt{2d-reacher} with obstacle. RRT* maintains a superior performance, but LAMPO achieves 0.93 success rate with expert demonstrations from RRT*. \textbf{Right}, an example of trajectory provided by RRT*.}
 	\hspace{-0.2cm}\input{figures/ablation2.tex}\\
 	%\hspace{-2em}
 	\begin{tikzpicture}

\definecolor{color16}{rgb}{0.75,0.75,0}
\definecolor{color17}{rgb}{1,0.25,0}
\definecolor{color18}{rgb}{0,0.75,0.75}
\definecolor{color19}{rgb}{0,0.25,1}

\begin{axis}[
height=2cm,
width=8cm,
hide axis,
xmin=10,
xmax=50,
ymin=0,
ymax=0.5,
legend columns=-1,
legend entries={{CT (IM)},{CT (RL)},{LAMPO (IM)}, {LAMPO (RL)}},
legend style={at={(0.7,1.0)}, anchor=north, draw=none, font=\scriptsize, column sep=1pt, line width=2 pt},
]

\addlegendimage{no markers, color16}
\addlegendimage{no markers, color17}
\addlegendimage{no markers, color18}
\addlegendimage{no markers, color19}

\end{axis}

\end{tikzpicture}\\
 	\vspace{-0.2cm}
 	\caption{\label{fig:samplesize}\footnotesize \texttt{reach-target}: (a) While keeping the number of samples per iteration fixed to $500$, we use different number of demonstrations. (b) While keeping the number of demonstrations fixed to $1000$, we use different amount of samples per policy improvement.}
 	\vspace{-0.6cm}
 \end{figure}
\begin{figure}
	\captionof{table}{\footnotesize Number of episodes required to achieve a desired success rate (S.R) performance.}\label{tab:deep}
\centering
\footnotesize
\adjustbox{width=0.48\textwidth}{%
	\begin{tabular}{l|| c| c| c || c | c | c} 
		\hline
		 & \multicolumn{3}{c||}{\texttt{reach-target}} & \multicolumn{3}{c}{\texttt{close-drawer}} \\
		\cline{2-7}
		\hline
		S.R & LAMPO & PPO & SAC & LAMPO & PPO & SAC \\ [0.5ex] 
		\hline\hline
		0.5 & \textbf{1000} & NaN & 10269 & \textbf{1000} & 5721 & 1947  \\
		\hline
		0.7 & \textbf{1600} & NaN & 13326  & \textbf{1000} & 8640 & 2404   \\
		\hline
		0.9 & NaN & NaN & \textbf{20369}  & NaN & 23402 & \textbf{3856}    \\
		\hline
	\end{tabular}}
\normalsize
\vspace{-0.7cm}
\end{figure}

%>>>>>>> 74392bb62d7a40871d2fe712272c76777ecfdd21
\noindent\texttt{reach-target:} This task, defined in the RLBench suite \cite{james_rlbench_2020}, comprises a 6dof robotic arm, equipped with a simple gripper. The goal is to reach an object in the scene (Fig.~\ref{fig:reacher2d}-center). Note that, to the best of our knowledge, we are the first to conduct experiments on the RLBench suite, hence there are no comparative results in literature. 
% Each demonstration is perturbed with a Gaussian noise with $\sigma^2=0.03$. The movement is encoded with $20$ radial basis functions, and since we need to encode $6$ dof, plus 2 for the gripper and an additional parameter for the movement's duration, we obtain primitives with $141$ parameters. 
% While the original reward is binary, we opted for a more informative reward, giving $1$ in case of success (object reached), or the negative euclidean distance to the goal in case of failure.

\noindent\texttt{close-drawer:} This task,  utilizes the same robot as above, and consists of closing a drawer, which appears in the scene in different positions and with different orientations (see Fig. \ref{fig:intro}-left). The context vector has $94$ dimensions. Note the high-dimensionality of this task, keeping in mind that RLBench provides us only with 200 demonstrations.
All reported results for the simulation environments have been averaged over 10 different experiments.

\noindent\textbf{{Real robot experiments}.}\\
\noindent\texttt{rr-reacher:} This task is learned and performed on a real $7$dof robot. The goal is to reach the position of a marker on a table (Fig.~\ref{fig:reacher2d}-right). 
% For the imitation phase, we recorded a dataset of $100$ demonstrations. 
% In order to learn the task, we sub-sample $50$ demonstrationsm for each run (we collected in totals 3 runs).

\noindent\texttt{rr-pouring:} On the same robot as before, we execute a task where the goal is to pour some granular material from a glass into a bowl, till the desired amount (in grams) of material is poured. Here, the context is 1d, i.e. the quantity of the material, measured by a digital scale placed under the bowl (see Fig. \ref{fig:intro}-right). This task, although lower-dimensional than the previous one, is harder to learn, as there is some stochasticity perturbing the experiments, e.g., small variations of the sugar quantity from episode to episode.

\noindent\textbf{Note:} The previously described reaching tasks can be handled by motion planning methods, and have been used for evaluating the ability of LAMPO to encode high-dimensional movements with variable contexts. A good performance by LAMPO in planning tasks, provides good evidence about the applicability of the proposed method for learning tasks that do not have accurate model description.

\color{black}

\subsection{Evaluation in Virtual Environments}
\begin{figure*}
	\hspace{-0.2cm}
	\includegraphics[width=1.0\textwidth]{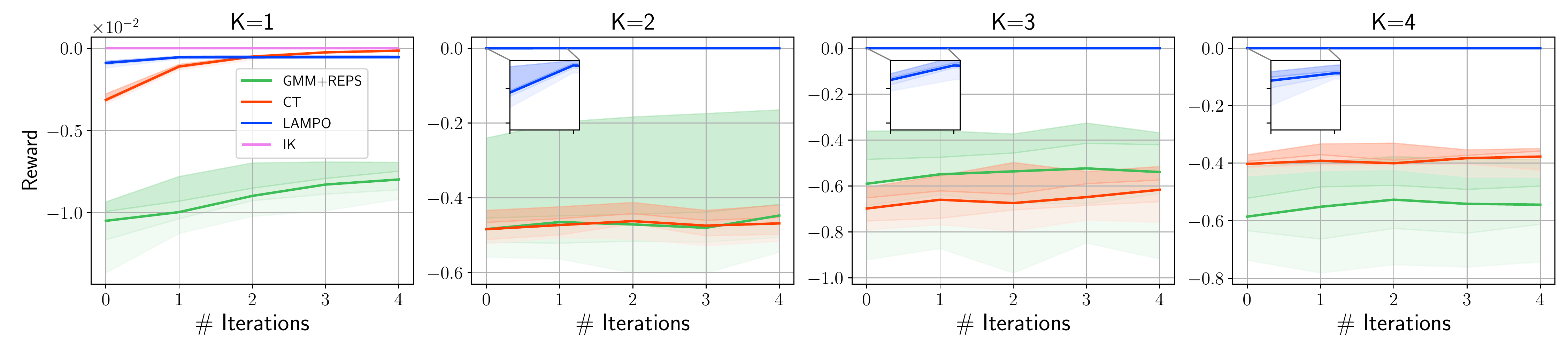}
	\vspace{-0.4cm}
	\caption{\label{fig:nonlinearrl}\footnotesize Policy improvement performance of LAMPO against the baseline method CT and  GMM+REPS in the \texttt{reacher-2d} task, using a different number of clusters $K$. CT struggles to learn at the presence of more than one clusters, while GMM+REPS performs poorly at all cases.}
	\vspace{-0.1cm}
	% This file was created by tikzplotlib v0.8.4.
\begin{tikzpicture}

\definecolor{color0}{rgb}{0.,0.25,1}

\begin{axis}[
tick align=outside,
ylabel={Success Rate},
ylabel style={font=\scriptsize},
xlabel={\# Iterations},
xlabel style={font=\scriptsize, yshift=6pt},
yticklabel style={font=\tiny},
xticklabel style={font=\tiny},
tick pos=left,
width=0.54\columnwidth,
grid=both,
title={(a) \texttt{reach-target}},
title style={yshift=-10pt},
x grid style={white!69.0196078431373!black},
xmin=-0.2, xmax=4.2,
xtick style={color=black},
y grid style={white!69.0196078431373!black},
ymin=0.56563147515405, ymax=0.785869976341395,
ytick style={color=black}
]
\addplot [semithick, color0]
table[row sep=\\] {%
0 0.6023\\
1 0.6539\\
2 0.6913\\
3 0.7265\\
4 0.7335\\
};
\path [draw=color0, semithick]
(axis cs:0,0.575642316117112)
--(axis cs:0,0.628957683882888);

\path [draw=color0, semithick]
(axis cs:1,0.626404498731611)
--(axis cs:1,0.681395501268389);

\path [draw=color0, semithick]
(axis cs:2,0.655577422965301)
--(axis cs:2,0.727022577034699);

\path [draw=color0, semithick]
(axis cs:3,0.68843189852383)
--(axis cs:3,0.76456810147617);

\path [draw=color0, semithick]
(axis cs:4,0.691140864621666)
--(axis cs:4,0.775859135378334);

\addplot [semithick, color0, mark=-, mark size=4, mark options={solid}, only marks]
table[row sep=\\] {%
0 0.575642316117112\\
1 0.626404498731611\\
2 0.655577422965301\\
3 0.68843189852383\\
4 0.691140864621666\\
};
\addplot [semithick, color0, mark=-, mark size=4, mark options={solid}, only marks]
table[row sep=\\] {%
0 0.628957683882888\\
1 0.681395501268389\\
2 0.727022577034699\\
3 0.76456810147617\\
4 0.775859135378334\\
};
\addplot [line width=2pt, color0]
table[row sep=\\] {%
0 0.6023\\
1 0.6539\\
2 0.6913\\
3 0.7265\\
4 0.7335\\
};
\end{axis}

\end{tikzpicture}
	% This file was created by tikzplotlib v0.8.4.
\begin{tikzpicture}

\definecolor{color0}{rgb}{0.,0.25,1}

\begin{axis}[
width=0.54\columnwidth,
yticklabel style={font=\tiny},
xticklabel style={font=\tiny},
xlabel={\# Iterations},
xlabel style={font=\scriptsize, yshift=6pt},
tick align=outside,
tick pos=left,
title={(b) \texttt{close-drawer}},
title style={yshift=-10pt},
x grid style={white!69.01960784313725!black},
y grid style={white!69.01960784313725!black},
grid=both,
xmin=-0.2, xmax=4.2,
xtick style={color=black},
y grid style={white!69.0196078431373!black},
ymin=0.71, ymax=0.83,
ytick style={color=black}
]
\addplot [semithick, color0]
table[row sep=\\] {%
	0 0.7348\\
	1 0.7504\\
	2 0.7696\\
	3 0.7876\\
	4 0.7898\\
};
\path [draw=color0, semithick]
(axis cs:0,0.713694429171425)
--(axis cs:0,0.755905570828575);

\path [draw=color0, semithick]
(axis cs:1,0.709650957311858)
--(axis cs:1,0.791149042688142);

\path [draw=color0, semithick]
(axis cs:2,0.737745209465451)
--(axis cs:2,0.801454790534549);

\path [draw=color0, semithick]
(axis cs:3,0.750767040846546)
--(axis cs:3,0.824432959153454);

\path [draw=color0, semithick]
(axis cs:4,0.759978351487552)
--(axis cs:4,0.819621648512448);

\addplot [semithick, color0]
table[row sep=\\] {%
	0 0.7348\\
	1 0.7504\\
	2 0.7696\\
	3 0.7876\\
	4 0.7898\\
};
\addplot [semithick, color0, mark=-, mark size=4, mark options={solid}, only marks]
table[row sep=\\] {%
	0 0.713694429171425\\
	1 0.709650957311858\\
	2 0.737745209465451\\
	3 0.750767040846546\\
	4 0.759978351487552\\
};
\addplot [semithick, color0, mark=-, mark size=4, mark options={solid}, only marks]
table[row sep=\\] {%
	0 0.755905570828575\\
	1 0.791149042688142\\
	2 0.801454790534549\\
	3 0.824432959153454\\
	4 0.819621648512448\\
};
\addplot [line width=2pt, color0]
table[row sep=\\] {%
	0 0.7348\\
	1 0.7504\\
	2 0.7696\\
	3 0.7876\\
	4 0.7898\\
};
\end{axis}

\end{tikzpicture}
	% This file was created by tikzplotlib v0.8.4.
\begin{tikzpicture}

\definecolor{color0}{rgb}{0.,0.25,1}

\begin{axis}[
title={{\tiny $10^{-3}$} (c) \texttt{rr-reacher}},
title style={yshift=-10pt, xshift=-10pt},
ylabel={Reward},
yticklabel style={font=\tiny},
xticklabel style={font=\tiny},
ylabel style={yshift=-10pt, font=\scriptsize},
xlabel={\# Iterations},
xlabel style={font=\scriptsize, yshift=6pt},
width=0.54\columnwidth,
tick align=outside,
tick pos=left,
x grid style={white!69.0196078431373!black},
xmajorgrids,
xmin=-0.2, xmax=4.2,
xtick style={color=black},
y grid style={white!69.0196078431373!black},
ymajorgrids,
ymin=-0.728139447605617, ymax=-0.204181279220089,
ytick style={color=black}
]
\addplot [semithick, color0]
table[row sep=\\] {%
0 -0.59955492302057\\
1 -0.375298490441481\\
2 -0.283229212723872\\
3 -0.301837573100165\\
4 -0.324435490481757\\
};
\path [draw=color0, semithick]
(axis cs:0,-0.704323167224456)
--(axis cs:0,-0.494786678816684);

\path [draw=color0, semithick]
(axis cs:1,-0.388765583708888)
--(axis cs:1,-0.361831397174075);

\path [draw=color0, semithick]
(axis cs:2,-0.338460865846495)
--(axis cs:2,-0.227997559601249);

\path [draw=color0, semithick]
(axis cs:3,-0.328538335593831)
--(axis cs:3,-0.275136810606499);

\path [draw=color0, semithick]
(axis cs:4,-0.338871194282244)
--(axis cs:4,-0.30999978668127);

\addplot [semithick, color0, mark=-, mark size=4, mark options={solid}, only marks]
table [row sep=\\]{%
0 -0.704323167224456\\
1 -0.388765583708888\\
2 -0.338460865846495\\
3 -0.328538335593831\\
4 -0.338871194282244\\
};
\addplot [semithick, color0, mark=-, mark size=4, mark options={solid}, only marks]
table [row sep=\\]{%
0 -0.494786678816684\\
1 -0.361831397174075\\
2 -0.227997559601249\\
3 -0.275136810606499\\
4 -0.30999978668127\\
};
\addplot [line width=2pt, color0]
table[row sep=\\] {%
0 -0.59955492302057\\
1 -0.375298490441481\\
2 -0.283229212723872\\
3 -0.301837573100165\\
4 -0.324435490481757\\
};
\end{axis};
%\node[anchor=east] at (title.west) {A};
\end{tikzpicture}
	% This file was created by tikzplotlib v0.8.4.
\begin{tikzpicture}

\definecolor{color0}{rgb}{0.,0.25,1}

\begin{axis}[
title={(d) \texttt{rr-pouring}},
title style={yshift=-10pt},
yticklabel style={font=\tiny},
xticklabel style={font=\tiny},
xlabel={\# Iterations},
xlabel style={font=\scriptsize, yshift=6pt},
width=0.54\columnwidth,
tick align=outside,
tick pos=left,
x grid style={white!69.0196078431373!black},
xmajorgrids,
xmin=-0.4, xmax=8.4,
xtick style={color=black},
y grid style={white!69.0196078431373!black},
ymajorgrids,
ymin=-0.682170028400198, ymax=-0.0528944594719905,
ytick style={color=black}
]
\addplot [semithick, color0]
table[row sep=\\] {%
0 -0.544\\
1 -0.3904\\
2 -0.448\\
3 -0.3264\\
4 -0.2712\\
5 -0.2088\\
6 -0.1512\\
7 -0.1304\\
8 -0.108\\
};
\path [draw=color0, semithick]
(axis cs:0,-0.653566593448916)
--(axis cs:0,-0.434433406551084);

\path [draw=color0, semithick]
(axis cs:1,-0.478300694513752)
--(axis cs:1,-0.302499305486248);

\path [draw=color0, semithick]
(axis cs:2,-0.537538192968141)
--(axis cs:2,-0.358461807031859);

\path [draw=color0, semithick]
(axis cs:3,-0.394604485013817)
--(axis cs:3,-0.258195514986183);

\path [draw=color0, semithick]
(axis cs:4,-0.317207565290939)
--(axis cs:4,-0.225192434709061);

\path [draw=color0, semithick]
(axis cs:5,-0.274813514252765)
--(axis cs:5,-0.142786485747235);

\path [draw=color0, semithick]
(axis cs:6,-0.195100378859413)
--(axis cs:6,-0.107299621140587);

\path [draw=color0, semithick]
(axis cs:7,-0.149997981936924)
--(axis cs:7,-0.110802018063076);

\path [draw=color0, semithick]
(axis cs:8,-0.134502105576727)
--(axis cs:8,-0.0814978944232727);

\addplot [semithick, color0, mark=-, mark size=4, mark options={solid}, only marks]
table[row sep=\\] {%
0 -0.653566593448916\\
1 -0.478300694513752\\
2 -0.537538192968141\\
3 -0.394604485013817\\
4 -0.317207565290939\\
5 -0.274813514252765\\
6 -0.195100378859413\\
7 -0.149997981936924\\
8 -0.134502105576727\\
};
\addplot [semithick, color0, mark=-, mark size=4, mark options={solid}, only marks]
table[row sep=\\] {%
0 -0.434433406551084\\
1 -0.302499305486248\\
2 -0.358461807031859\\
3 -0.258195514986183\\
4 -0.225192434709061\\
5 -0.142786485747235\\
6 -0.107299621140587\\
7 -0.110802018063076\\
8 -0.0814978944232727\\
};
\addplot [line width=2pt, color0]
table[row sep=\\] {%
0 -0.544\\
1 -0.3904\\
2 -0.448\\
3 -0.3264\\
4 -0.2712\\
5 -0.2088\\
6 -0.1512\\
7 -0.1304\\
8 -0.108\\
};
\end{axis}

\end{tikzpicture}% \\
	\vspace{-0.25cm}
	\caption{\label{fig:learning-robot-reach}\footnotesize Learning curve accompanied with $95\%$ confidence intervals on simulated and real robotic environments. LAMPO shows often a substantial improvement over the imitation-learning phase.}
	\vspace{-0.8cm}
\end{figure*}

For providing insights regarding the performance of LAMPO, we have compared it against the most relevant algorithm by Colom\'e and Torras (CT) \cite{colome_dimensionality_2018-1}, that uses GMMs in the parameter space of ProMPs only after performing DR, and optimizes the policy using REPS, but also with IK-based and RRT* planning. 
%Specifically, due to the assumptions done by \cite{colome_dimensionality_2018-1} in their algorithmic solution, 
We decided, first, to compare on the toy experiment of \texttt{2d-reacher} 
%First, we evaluate the selection of different number of clusters and latent space dimensions used for the MMPCA and GMMs approaches in LAMPO and CT respectively, in imitation learning. Fig. \ref{fig:nonlinear} shows the heatmaps of the ablation study conducted over the two algorithms, for different numbers of clusters and latent dimensions. Note that the result depicts the performance of the obtained rewards in \texttt{2d-reacher}, i.e. Euclidean distance to the goal. It is evident that CT algorithm cannot decode different clusters of motions when compared to our approach. 
for the policy improvement setting, depicted in Fig. \ref{fig:nonlinearrl}. In this evaluation, we keep the latent space dimension fixed for CT and LAMPO, but we test the performance for different number of goal-clusters (from 1 to 4). For fair comparison, we have also included another baseline method, GMM+REPS, in which we do not appy DR on the collected data. As expected, while for unimodal movements ($K=1$) LAMPO and CT have similar performance, for multimodal movements ($K>1$), CT performance is suboptimal. The simple GMM+REPS approach performs poorly in all cases, showing the importance of acquiring a low-dimensional latent representation that preserves the motion's non-linearities. 
%Evidently, as IK provides optimal solution, LAMPO achieves same performance when employing multiple movement clusters. 
Furthermore, we tested LAMPO on a more challengind setting of the \texttt{2d-reacher} with an obstacle, and compared with RTT*, as depicted in Fig.~\ref{fig:obstacle}.
% Notably, LAMPO's expert demonstrations come from RRT*'s possible paths.
LAMPO achieves a good performance of 0.93 success rate of reaching the target goal positions. We argue that our method can scale quite well in high-dimensional tasks, which allows us to trade-off the agent's performance with the possibility of applying our algorithm when there is not an accurate description of the model or the task's objective.

After the above observations, we evaluated LAMPO on high dimensional problems. Starting with the \texttt{reach-target} task, we ablated the performance of LAMPO and CT for the RL setting (Fig. \ref{fig:samplesize}). In Fig.~\ref{fig:samplesize}a, we compare both algorithms for different initial number of demonstrations in the imitation learning phase. The light-green and light--blue curves indicate performance for the IM policy, while the red and blue the curves for the RL policy for CT and LAMPO respectively.
We tested the influence of different batch sizes during the RL phase, while using the same number of samples for imitation learning (Fig.~\ref{fig:samplesize}b). LAMPO outperforms the baseline in both cases, when CT struggles with the multimodality of the movements in the respective task.
After collecting only $200$ samples during RL, the performance of LAMPO stabilizes, meaning that the collection of more samples is unnecessary,  empirically proving the sample-efficiency of LAMPO.

%\begin{figure}
%	\input{figures/reacher-obstacle-plot.tex}
%	\begin{subfigure}{0.39\columnwidth}
%		\vspace{-5.38cm}
%	\input{figures/obstacle-env.tex}
%	\end{subfigure}
%	% \input{figures/legend1.tex}
%	\caption{\label{fig:obstacle}\small On the left, learning curve of LAMPO in the \texttt{reacher2d-obstacle} environment. RRT* maintains a superior performance. On the right, an example of trajectory provided by RRT*.}
%	\vspace{-0.5cm}
%\end{figure}
%From the results of the previous comparative study, we have concluded on a hyper-parameter setting for LAMPO, which can provide improved policies in a data-efficient way. 
For further evaluation, we also conducted experiments of LAMPO in another simulated high-dimensional problem, the \texttt{close-drawer} (Fig.~\ref{fig:learning-robot-reach}b). 
%While we do not achieve the same success rate as in \texttt{reach-target} in Fig.~\ref{fig:learning-robot-reach}a, we can still see how 
Arguably, LAMPO ameliorates the initial poor performance obtained by IL. In the future, we want to study more expressive models for learning MPs combined with LAMPO, to cover the whole space of solutions. 
We also compared LAMPO w.r.t. PPO and SAC on the \texttt{reach-target} and \texttt{close-drawer}. From the results depicted in Table~\ref{tab:deep}, we deduce that LAMPO exhibits a higher sample efficiency w.r.t. state-of-the-art deep-RL in these tasks.
\subsection{Validation on real robot}
To further validate the efficieny of LAMPO, we tested it on a real robotic manipulator. %As we have already described in subsection \ref{sec:details}, 
We have trained and validated LAMPO on two tasks. Fig. \ref{fig:learning-robot-reach}c depicts the results of learning with LAMPO the \texttt{rr-reacher} task, where we can observe the consistent improvement during the RL phase. Indeed, we end up with a policy that achieves significantly lower error. 
% Note that there is also a consistent error in the measurement of the fingertip and the target position from the motion capture system, e.g. due to possible occlusions or the nominal system noise.

Fig. \ref{fig:learning-robot-reach}d shows the results obtained for the \texttt{rr-pouring} task. We can see an impressive slope in the learning curve, that starts from a low performance after imitation learning, achieving optimal results after improving the policy with LAMPO. Note that, also, in this case the results are dependent on the digital scale nominal error ($\sim 2$grams) and other sources of noise. 
When the environment's dynamics are difficult to capture (e.g., fluid's dynamics) planning techniques potentially lead to failure. DeepRL tends to be inefficient, and its initial explorative behavior can damage the hardware. LAMPO provides a safe and sample efficient way for skill learning, combining IL in a latent multimodal space, while not allowing the agent to forget the good, safe demonstrated policies while improving through the RL exploration.

% \section{Empirical Analysis}

% \subsection{Experimental Detail and Tasks}

% \subsection{An Ablation Study}

% \subsection{Validation on Virtual Environements}

% \subsection{Algorithm Validation on The Real Robot}

%%%%%%%%%%%%%%%%%%%%%%%%%%%%%%%%%%%%%%%%%%
\section{Conclusions \& Future work}
We introduced LAMPO, a novel contextual RL algorithm for optimizing over MPs while learning high-dimensional robotic manipulation tasks with different contexts in a safe way. The main advantage of LAMPO is twofold; first, it offers a reduced latent representation of non-linear dynamics, while encoding the dependencies between movements and task-related contexts introducing the MPPCA graphical model on MPs. Second, it gains in sample-efficiency with off-policy gradient estimations, re-using samples collected from previous learning episodes. 
%For learning the latent representation we u, which is fully probabilistic, and allows us to perform conditioning on the task contexts. For the off-policy estimation, we use SNIS combined with a full-gradient estimate, also considering the normalization factor, which further lowers the estimation variance. 
%Our contextual RL method uses a forward KL-divergence on the policy improvement, which, jointly with a KL-regularization of the context distribution, provides a robust policy optimization, \color{red} acting as safety constraint for not allowing the agent to forget demonstrations, gut gradually improve its policy\color{black} . 
We evaluated our algorithm both on simulated environments and on a real manipulator robot. The experimental analyses show that LAMPO outperforms the state of the art techniques in terms of sample efficiency on high-dimensional robotic manipulation tasks. Our future work will focus on introducing a decomposition of demonstrated tasks into sub-policies, to improve modularity and task performance.

\section*{Acknowledgment}
The research is funded by the BoschForschungsstiftung program. Dr. Chalvatzaki is funded by the DFG Emmy Noether Program (CH 2676/1-1). 
We would like to thank Oleg Arenz for the useful discussions.
%In the future, we will explore the possibility of introducing a Dirichlet prior to the mixture's weight in order to stabilize the learning, as well as a modification to the EM algorithm to incorporate the effect on the configuration space.  

%%%%%%%%%%%%%%%%%%%%%%%%%%%%%%%%%%%%%%%%%%%%%%%%%%%%%%%%%%%%%%%%%%%%%%%%%%%%%%%%
%\section*{APPENDIX}

% \section*{ACKNOWLEDGMENT}

\clearpage
\bibliographystyle{IEEEtran}
\bibliography{zotero}

\end{document}